%% file: main.tex
\documentclass[10pt,letterpaper]{article}

\usepackage{cogsci}

\cogscifinalcopy 

\usepackage{pslatex}
\usepackage{apacite}
\usepackage{float} 
\usepackage{amsmath}
\usepackage{graphicx}
\usepackage{subfigure}

\title{Selectively Providing Reliance Calibration Cues With Reliance Prediction}
 
\author{{\large \bf Yosuke Fukuchi (fukuchi@nii.ac.jp)} \\
  National Institute of Informatics, Tokyo, Japan
  \AND {\large \bf Seiji Yamada (seiji@nii.ac.jp)} \\
  National Institute of Informatics, Tokyo, Japan \\
  The Graduate University for Advanced Studies, SOKENDAI, Tokyo, Japan
  }

\begin{document}

\maketitle
\input{0_abstract}
\input{1_introduction}

\input{2_background}

\input{3_proposal}
\input{4_training}
\input{5_experiment}
\input{6_conclusion}

\bibliographystyle{apacite}

\setlength{\bibleftmargin}{.125in}
\setlength{\bibindent}{-\bibleftmargin}

\bibliography{bib}

\end{document}

%% file: 0_abstract.tex
\begin{abstract}
For effective collaboration between humans and intelligent agents that employ machine learning for decision-making,
humans must understand what agents can and cannot do to avoid over/under-reliance.
A solution to this problem is adjusting human reliance through communication using reliance calibration cues (RCCs) to help humans assess agents' capabilities.
Previous studies typically attempted to calibrate reliance by continuously presenting RCCs,
and when an agent should provide RCCs remains an open question.
To answer this, we propose Pred-RC, a method for selectively providing RCCs.
Pred-RC uses a cognitive reliance model to predict whether a human will assign a task to an agent.
By comparing the prediction results for both cases with and without an RCC, Pred-RC evaluates the influence of the RCC on human reliance.
We tested Pred-RC in a human-AI collaboration task
and found that it can successfully calibrate human reliance with a reduced number of RCCs.

\end{abstract}

%% file: 1_introduction.tex
\section{Introduction}
Machine learning (ML) is a powerful tool for robots and agents that collaborate with humans.
There have been many trials of introducing ML, and such AI agents have shown great performance in various fields~\cite{doi:10.1142/S0219622019300052,doi:10.1080/23311916.2019.1632046,pmlr-v87-kalashnikov18a}.
However, as ML models get more complex, it becomes more difficult for end users to understand how to adequately use AI agents~\cite{8466590,rai2020explainable},
a consequence of which is that users over-rely or under-rely on them~\cite{0cd2e253a4034377a227c5d5e7071e62,doi:10.1518/001872097778543886}. 
Over-reliance, in which a human overestimates the capability of an AI agent, can cause misuse and task failure~\cite{7451740}.
It even leads to even serious accidents, particularly for embodied agents such as robots and autonomous vehicles.
Under-reliance is also problematic because it results in disuse, increases human workload, and degrades the total collaboration performance.

Previous studies attempted to adjust human reliance
by providing signals or information elements used by humans to assess an AI's capability~\cite{10.1007/978-3-319-07458-0_24}.
In this paper, we call them reliance calibration cues (RCCs).
For example, presenting an AI model's confidence rate is shown to be effective for an RCC~\cite{mcguirl,10.1145/3351095.3372852}.

A challenge facing reliance calibration with RCCs lies in the timing at which to provide them.
In typical previous studies, RCCs are provided continuously,
but this is not always realistic,
for example when a robot verbally provides them.
There is a trade-off between successful calibration and reducing the communication cost,
but computational methods for deciding when to provide them are quite limited.
Okamura \& Yamada proposed a method for selectively providing RCCs by detecting over/under-reliance \cite{10.1371/journal.pone.0229132,9281021}.
However, their method simply checks a human's false assignment of past tasks to him/herself or an AI
and does not capture a human's cognitive aspects such as his/her past experiences collaborating with AI and beliefs of what kind of tasks the AI can do.

\begin{figure}
  \includegraphics[width=\linewidth]{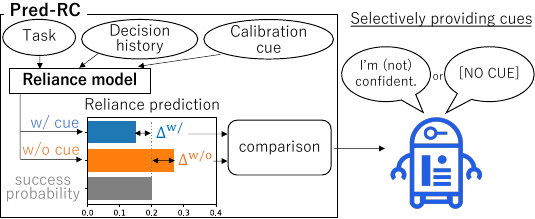}
  \caption{
    Predictive Reliance Calibrator (Pred-RC) enables AI system to selectively provide trust calibration cues.
    In Pred-RC, cognitive reliance model predicts probability that human will assign current task to AI.
    By considering both cases with and without cue provision, Pred-RC evaluates how much cue will contribute to trust calibration and decides whether to provide it.
  }
  \label{fig:teaser}
\end{figure}
This paper proposes {\it Predictive Reliance Calibrator (Pred-RC)}, a method for selectively providing RCCs (Fig. \ref{fig:teaser}).
The main idea of Pred-RC is that it selects whether to provide an RCC to avoid a wide gap between the human reliance rate and the AI's success probability.
Here, the reliance rate is the probability that a human will assign a current task to the AI system.
In Pred-RC, a cognitive reliance model predicts human reliance rates in both cases where an RCC is provided or not.
By comparing the predicted rates with the success probability (actual reliability), Pred-RC evaluates the impact of an RCC for reliance calibration.

This paper reports an experiment on reliance calibration by Pred-RC using crowdsourcing.
We focused on crowdworkers' decision accuracy, or how many times the workers assigned tasks that an AI could solve to the AI and did ones that the AI could not by themselves.
The results show that the workers' accuracy did not decline with Pred-RC's selective RCCs, whereas that of workers whose RCCs were randomly provided got worse 
as fewer RCCs were provided,
suggesting that Pred-RC enables an AI to selectively provide RCCs at the proper timing by predicting and comparing reliance with and without an RCC.

%% file: 2_background.tex
\section{Background}\label{section:background}

\subsection{Trust/reliance calibration}
Reliance is a concept relevant to trust, and it is sometimes studied inclusively in the field of engineering.
Trust is attitudinal and a psychological construct, while reliance focuses on the behaviors of humans,
which is directly observable and thus an objective measure~\cite{https://doi.org/10.48550/arxiv.2203.12318}.
Although the main focus of this paper is reliance, this section reviews both trust and reliance calibration to highlight our research
because of their close relevance.

There are various approaches to achieving trust/reliance calibration,
one of which is to change an agent's actions~\cite{Dubois2020AdaptiveTA,10.1145/3450267.3450529}.
For example, Chen et al. proposed trust-POMDP, a computational model that allows an AI to decide an action
with awareness of human trust~\cite{10.1145/3359616}.
They demonstrated that with trust-POMDP, a robot automatically generates behavior that involves
tackling an easier task first and successfully earns human trust.

Another approach is to explicitly provide information or communicative signals that help humans assess an AI's capability.
Commonly used RCCs are the confidence rate or uncertainty of an AI's decision-making~\cite{mcguirl,10.1145/3351095.3372852,10.1145/2516540.2516554}.
In this paper, Pred-RC provides confidence information as an RCC.

Some studies focused on continuously providing RCCs and demonstrated its effectiveness~\cite{10.1145/2516540.2516554,10.1145/3025171.3025198}.
McGuirl \& Sarter compared providing dynamic system confidence with overall reliability only
and found that the former can improve trust calibration~\cite{mcguirl}.

However, there are at least two potential concerns with the continuous provision of RCCs.
One is that too many RCCs can be annoying for humans.
This depends mainly on how the RCC are provided, 
and a typical negative case is that in which a robot verbally provides them.
The other concern is that humans sometimes pay less attention to continuously displayed information.
Okamura \& Yamada found that, in their experiment, participants did not change their over-reliance in spite of
a reliability indicator being continuously displayed to them. They also found that giving additional trigger signals was effective in resolving this problem~\cite{10.1371/journal.pone.0229132}.
Therefore, we aim to achieve successful reliance calibration by not continuously but selectively providing RCCs.

Very few studies have focused on a computational method for adaptively providing RCCs.
A method proposed by Okamura \& Yamada
judges whether an AI should provide an RCC or not with ``trust equations,'' logical formulae that mathematically express a human's over/under-reliance~\cite{10.1371/journal.pone.0229132,9281021}. 
A problem with this method is that its judgment depends only on how many times a human falsely assigns a task to an AI or him/herself,
and it cannot capture the details of collaboration experiences such as in what task a human observed an AI's failure, when an AI provided RCCs, and what the tasks were.
These experiences can affect human beliefs about an AI's capability.
For example, an experience with an AI's success/failure on a task is more likely to influence human reliance in a similar task than a different task.
In Pred-RC, a reliance model is trained to predict human reliance, taking into account the collaboration history between a human and an AI, and it is expected to capture these aspects.

\subsection{Reliance estimation}
A basic idea of Pred-RC is that inferring human reliance on an AI agent helps with the selective provision of RCCs.
For example, an RCC that increases human reliance may be less effective
if a human already has high reliance on an agent than if s/he has low reliance.

Self-report trust scales are commonly used to measure trust~\cite{doi:10.1207/S15327566IJCE040104,Madsen00measuringhuman-computer,Yagoda2012YouWM}.
Some studies focus on neural metrics to infer human trust using fMRI or EEG~\cite{10.1093/scan/nss144,9687123}.
A weak point of these methods is that they can be intrusive during a task execution.
A more relevant approach to this study focuses on human behavior.
Walker et al. proposed a method for inferring human trust on the basis of their gaze movements~\cite{Walker}.
Human interventions or takeovers of a robot's action is an indicator of poor trust,
and Muir incorporated them into human trust models~\cite{intervention}.
Another factor is a human's decision-making regarding whether to assign a task to themselves or an AI~\cite{10.1371/journal.pone.0229132,9281021},
and we follow this approach.

Many methods have been proposed to estimate reliance/trust,
but none of them can take into account the effects of RCCs on human reliance, or the effects of what RCCs have been provided so far and how the reliance changes if or unless an RCC is provided for a current task, which the reliance model aims to achieve.

%% file: 3_proposal.tex
\section{Selectively providing trust calibration cues}\label{section:proposal}
\subsection{Formalization}\label{section:formalization}
This paper formalizes human-AI collaboration with selectively provided RCCs as a tuple $(x, \hat{c}, c, d, y^*, y,  p)$.
Let us consider a situation in which a human sequentially performs a set of tasks $\{x_i\}^N_{i=1}$ with an AI agent,
where $i$ is the index of a task and $N$ is the number of tasks.
$\hat{c_i}$ is a potential RCC for the AI system when $x_i$ is given, and Pred-RC decides whether to provide it.
$c_i$ is the RCC actually provided to the human:
\begin{equation}\label{eqn:which-agent}
c_i =
  \begin{cases}
    \hat{c_i} & \mathrm{(if\ RCC\ is\ provided)} \\
    [MASK] & \mathrm{(elsewise)}.
  \end{cases}
\end{equation}
$c_i = [MASK]$ means that no RCC is provided.

The human observes $(x_i, c_i, y^*_i)$ and determines whether to assign $x_i$ to him/herself or the AI agent.
Let $d_i \in \{\mathrm{AI}, \mathrm{human}\}$ be the agent responsible for $x_i$.
$y^*_i$ is the desired result for $x_i$, and $y_i$ is the actual result for $x_i$ performed by $d_i$.
$y^*_i = y_i$ indicates the success of $x_i$.
The human can observe the result produced by the AI when $d_i = \mathrm{AI}$,
which is feedback for him/her to assess its reliability,
but cannot when $d_i = \mathrm{human}$.

$p_i$ is the success probability of the AI for $x_i$.
$i$ is incremented when $x_i$ is completed.

\subsection{Pred-RC}\label{ss:pred-tc}
Pred-RC adaptively selects whether to provide an RCC (Fig. \ref{fig:teaser}).
The main idea of Pred-RC is that it aims to avoid a discrepancy between the human reliance rate $r_i$ and the AI's success probability $p_i$.
Here, $r_i$ is the probability that the human will assign $x_i$ to the AI.
Pred-RC considers two types of $r_i$:
\begin{eqnarray}\label{eqn:r}
  r^{\mathrm{w/}}_i &=& P(d_i = \mathrm{AI}| x_{:i}, c_{:i-1}, c_{i}=\hat{c_i}, d_{:i-1}, y^*_{:i}, y_{:i-1}), \nonumber \\
  r^{\mathrm{w/o}}_i &=& P(d_i = \mathrm{AI}| x_{:i}, c_{:i-1}, c_{i}=[MASK], d_{:i-1}, y^*_{:i}, y_{:i-1}). \nonumber \\
\end{eqnarray}
The difference between $r^{\mathrm{w/}}$ and $r^{\mathrm{w/o}}$ is whether $\hat{c_i}$ is provided to the human or not.
Variables with the subscript $*_{:i}$ represent the vector of the sequence $(*_1, *_2, .., *_i)$.

The discrepancy $\Delta_i$ is the difference between $r_i$ and $p_i$:
\begin{equation}
  \begin{cases}
  \Delta^{\mathrm{w/}}_i &= |r^{w/}_i - p_i|, \\
  \Delta^{\mathrm{w/o}}_i &= |r^{w/o}_i - p_i|.
  \end{cases}
\end{equation}

Pred-RC compares $\Delta^{\mathrm{w/}}_i$ and $\Delta^{\mathrm{w/o}}_i$ and decides whether to provide $\hat{c_i}$.
Equation~\ref{eqn:which-agent} is rewritten as follows:
\begin{equation}\label{eqn:which-agent2}
c_i =
  \begin{cases}
    \hat{c_i} & (\Delta^{\mathrm{w/o}}_i - \Delta^{\mathrm{w/}}_i < threshold) \\
    [MASK] & \mathrm{(elsewise)}.
  \end{cases}
\end{equation}
$threshold$ represents the allowable range of $\Delta^{\mathrm{w/o}}_i$ compared with $\Delta^{\mathrm{w/}}_i$ and controls how much Pred-RC omits RCCs.
$threshold = 0$ means that Pred-RC omits $c_i$ only when no RCC is predicted to be better rather than providing $\hat{c_i}$,
and increasing $threshold$ results in more omitted RCCs.

\subsection{Reliance model}
\begin{figure}
  \includegraphics[width=\linewidth]{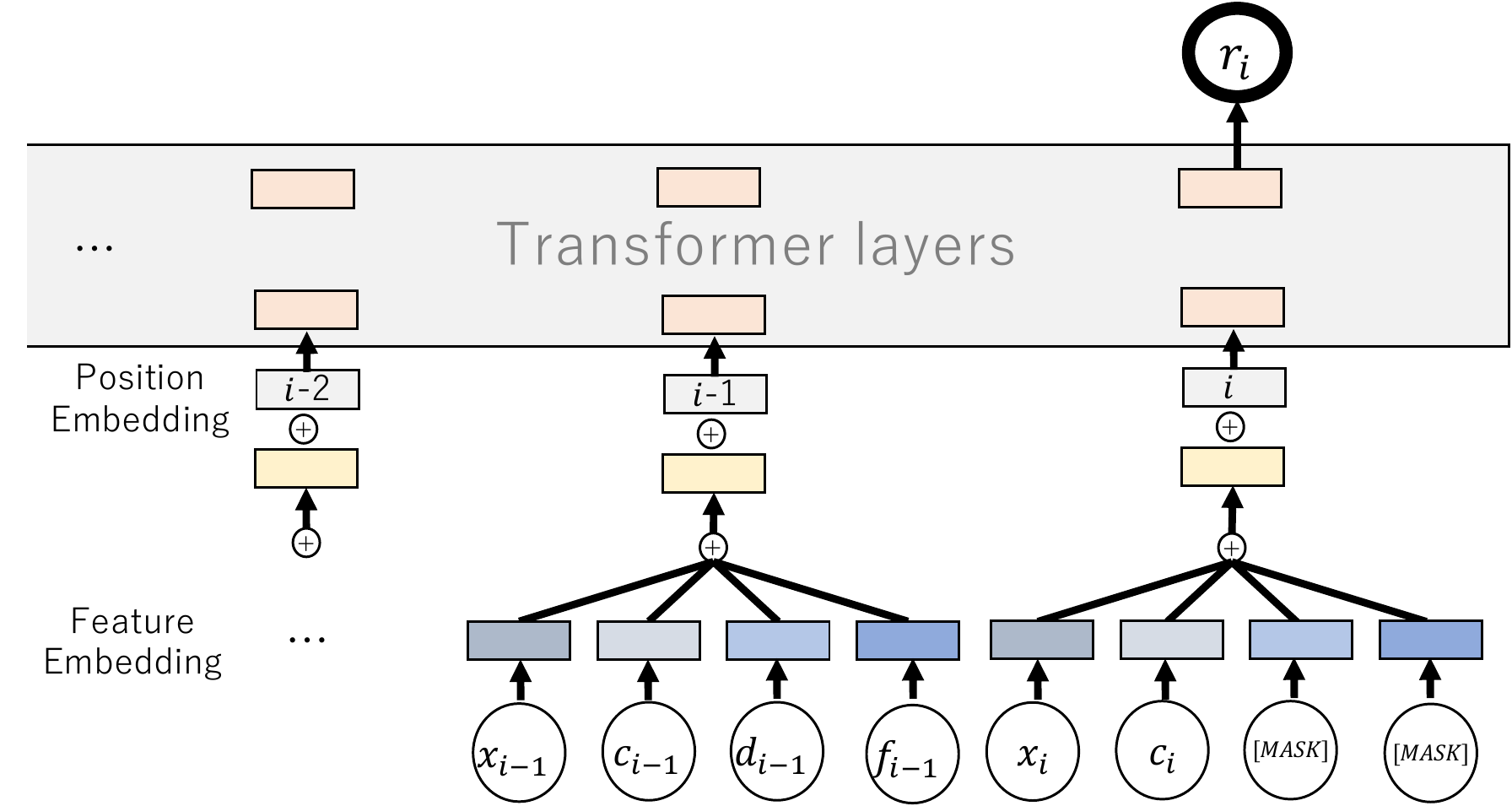}
  \caption{Structure of reliance model}
  \label{fig:trust-former}
\end{figure}
The reliance model is a cognitive model that predicts both $r_i^{\mathrm{w/}}$ and $r_i^{\mathrm{w/o}}$.
Figure~\ref{fig:trust-former} illustrates the structure of the model.
It is based on the Transformer encoder~\cite{NIPS2017_3f5ee243}, a deep-learning model that has shown great performance originally in natural language processing and 
is being applied to various domains~\cite{9716741,earthquake} including human-computer interaction~\cite{Matsumori_2021_ICCV,9777698}.
By taking into account the collaboration history between a human and the target AI system,
the reliance model can effectively capture human beliefs regarding an AI's capability.

The reliance model receives a history of collaboration between a human and AI ($x_{:i-1}, c_{:i-1}, d_{:i-1}, f_{:i-1}$) and the current state ($x_i, c_i$).
The history includes information such as when and to which task an RCC was provided and which decision the human made regarding the task,
so the reliance model can capture a human's beliefs regarding what task they think the AI can execute to predict human decisions better.

Each feature in the collaboration history is first embedded with perceptrons.
The embedded vectors are summed up with position embeddings, which give index information~\cite{NIPS2017_3f5ee243}.
Then, the vectors are transformed by the Transformer encoder model,
and a multi-layer perceptron predicts $r_i$ from the transformed vector of the index $i$.

Unlike equation~\ref{eqn:r}, the reliance model cannot access $y^*$ because we assume that the AI is not perfect,
although it is an important information for the human to assess the AI's reliability by comparing it with the AI's result.
Instead of this, we included $f$, feedback from the human's task result:
\begin{equation}\label{eqn:f}
f_i =
  \begin{cases}
    0 & (d_i = \mathrm{AI}) \\
    1 & (d_i = \mathrm{human\ and\ } y_i \mathrm{\ matches\ the\ AI's\ result}) \\
    2 & (d_i = \mathrm{human\ and\ } y_i \mathrm{\ does\ not\ match\ the\ AI's\ result}).
  \end{cases}
\end{equation}
We masked $d_i$ and $f_i$ because they are not obtained when predicting $r_i$.

The reliance model is trained in a supervised manner.
We adopted a binary cross-entropy loss function for the training:
\begin{equation}\label{eqn:loss}
  L = -\delta(d_i, \mathrm{AI}) \cdot \mathrm{log}(r_i) - \delta(d_i, \mathrm{human}) \cdot \mathrm{log}(1 - r_i),
\end{equation}
where $\delta(a, b) = 1$ when $a = b$ and 0 when $a \neq b$.

When inferring $r_i$, we run the reliance model with both cases in which $c_t = \hat{c_t}$ and $c_t = [MASK]$,
the results of which are the predicted values of $r^\mathrm{w/}$ and $r^\mathrm{w/o}$, respectively.

%% file: 4_training.tex
\section{Training reliance model}\label{section:training}
\subsection{Task}\label{ss:task}
\begin{figure}[t]
  \begin{center}
  \includegraphics[width=.9\linewidth]{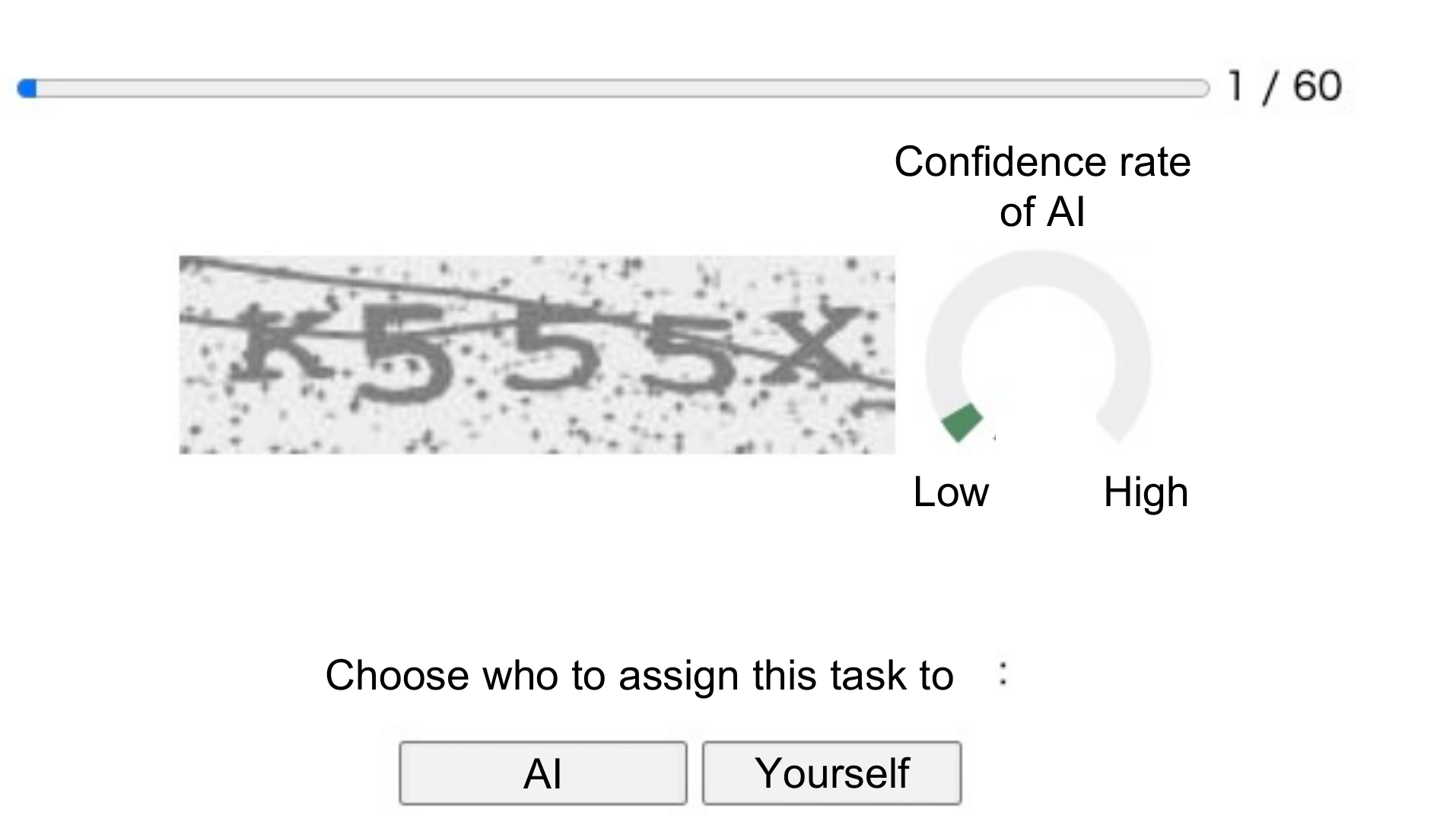}
    \caption{Screenshot of user interface for CC task}
  \label{fig:ui}
  \end{center}
\end{figure}
We developed a collaborative CAPTCHA (CC) task for training the reliance model and evaluating Pred-RC.
Figure~\ref{fig:ui} shows a screenshot of the user interface.
CAPTCHA is originally a task in which a human enters characters written in a noised and distorted image~\cite{10.1007/3-540-39200-9_18}.
In the CC task, a worker can get assistance from a task AI that is trained to recognize characters in images.
Here, $x_i$ is an image, and $y^*_i$ is a ground-truth label for $x_i$.
$\hat{c_i}$ is the confidence rate of the task AI for $x_i$.

A worker first chooses $d_i$ (Fig.~\ref{fig:ui}).
If s/he chooses ``AI,'' the task AI automatically enters its answer in a text box.
The worker can observe the AI's answer before sending it to the host server but cannot edit it.
If s/he chooses ``Yourself,'' an empty text box appears, and s/he is asked to enter the characters.
The worker repeats this 60 times.

\subsection{Task implementation}
\subsubsection{CAPTCHA dataset and task AI}
\begin{figure}[t]
  \begin{center}
    \subfigure[90.0\%]{\includegraphics[width=0.41\linewidth]{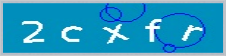}}
    \subfigure[59.8\%]{\includegraphics[width=0.41\linewidth]{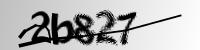}}
    \subfigure[0.0\%]{\includegraphics[width=0.38\linewidth]{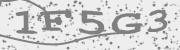}}
    \subfigure[0.0\%]{\includegraphics[width=0.28\linewidth]{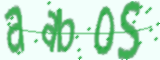}}
    \caption{Examples from CAPTCHA datasets. We used upper two datasets for training of task AI. Sub-caption shows accuracy of task AI for each dataset.}
  \label{fig:captcha}
  \end{center}
\end{figure}
Figure~\ref{fig:captcha} shows examples of CAPTCHA images used in our experiments.
We acquired four datasets from Kaggle, a web platform for data scientists and machine learning practitioners\footnote{https://www.kaggle.com/.}.
We split each dataset for training and testing.
We excluded two datasets for training the task AI to replicate a bias in AI capability.
For workers, understanding bias can help improve task assignment and result in fewer RCC requirements.
Figure~\ref{fig:captcha} also shows the accuracy of the task AI.
The accuracy is actually biased by the dataset used for the training.

Each CAPTCHA image has five characters, and the task AI outputs the probability distribution that the $j$-th character $x_{i, j}$ is $\iota \in I$,
where $I$ is a set of alphabetic and numerical characters.
\begin{equation}
  \mathrm{TaskAI}(x_{i, j}, \iota) = P(x_{i, j} = \iota).
\end{equation}
When $d_i =$ AI, $y_i$ is a sequence of the most probable $\iota \in I$ for each $x_{i, j}$.
\begin{equation}
  y^{\mathrm{AI}}_i = \{\mathrm{argmax}_\iota ( \mathrm{TaskAI}(x_{i, j}, \iota))\}_{j=1}^5.
\end{equation}
The task AI was implemented using ResNet-18, a deep-learning model commonly used for image processing.

\subsubsection{RCC and success probability}
The confidence rate was calculated on the basis of the probability distribution output from the task AI~\cite{pmlr-v70-guo17a}.
\begin{equation}
  \hat{c_i} \propto \Pi_{j=1}^5 (\mathrm{max}_{\iota \in I} (\mathrm{TaskAI}(x_{i, j}, \iota))).
\end{equation}
$\hat{c_i}$ becomes higher the more probability there is that the task AI assigns to the most probable character.
$p_i$ was calculated on the basis of $\hat{c_i}$ using logistic regression.
The logistic regression model was trained to predict whether $y^{\mathrm{AI}}_i$ matches $y^*_i$ from training datasets.

We adopted the confidence rate, $\hat{c_i}$, rather than the success probability because, in our pilot experiment,
we found that the confidence rate calibrates worker reliance better than the success probability.
This is presumably because the success probability, which was calculated with logistic regression, was distributed steeply around 0\% and 100\%,
whereas the distribution of the confidence rate was flatter. 

\subsection{Reliance dataset acquisition and model training}\label{ss:acquisition}
We made a {\it reliance dataset} to train the reliance model and evaluate Pred-RC.
The dataset was composed of sequences of the tuple $(x, \hat{c}, c, d, y^*, y,  p)$.

250 participants were recruited with compensation of 100 JPY through Yahoo! Japan crowdsourcing\footnote{https://crowdsourcing.yahoo.co.jp/}.
The data acquisition was conducted on a website.
The participants were first provided pertinent information, and all participants consented to the participation.
We instructed them on the flow of the CC task and asked five questions to check their comprehension of the task.
99 participants, who failed to answer the questions correctly, were excluded from the CC task,
and 151 participants remained (60 female and 91 male; aged $20-78, M=47.5, SD=12.2$).
The protocol of the reliance dataset acquisition and the evaluation of Pred-RC was approved by \{an anonymized ethics committee\}.

$x_i$ was randomly chosen for each participant from the test sub-datasets.
We manipulated how many images to use from each CAPTCHA dataset so that the task AI's overall accuracy became 50\% 
while keeping the task AI's accuracy for each dataset the same
to avoid extreme over/under-reliance.

Whether to provide an RCC was randomly decided for each participant.
The percentage of times that RCCs were provided was controlled to be 0, 20, 40, 60, 80, or 100\%. 

We trained the reliance model and investigated its accuracy with the reliance dataset.
We performed k-fold cross validation with stratification of the data to align the percentage of the number of provided RCCs.
We set $k=10$.
After 50 epochs of training, the reliance model predicted $d_i$ with a maximum accuracy of 81.6\% (95\% CI\footnote{Confidence interval}: 80.0\%, 83.2\%) on average at the 25th epoch.


%% file: 5_experiment.tex
\section{Evaluation}\label{section:experiment}
\subsection{Aim}
We evaluated whether Pred-RC can selectively provide RCCs at an effective timing.
More specifically, we investigated whether Pred-RC can reduce the number of RCCs while avoiding over/under-reliance.

\subsection{Procedure}
The CC task was used to evaluate Pred-RC.
The participants performed the task in a similar same way as the reliance dataset acquisition.
The difference is that it was Pred-RC that determined whether to provide RCCs with each participant's decision-making history, whereas this was randomly determined in the reliance dataset acquisition.
Pred-RC predicted the user reliance rate with the reliance model.

91 crowdworkers, none of whom participated in the data acquisition for the reliance dataset, were recruited for this experiment with compensation of 100 JPY.
Using the comprehension checking questions, 39 participants were excluded from the CC task,
and 52 participants remained (14 female and 38 male; aged 21-65, M=43.3, SD=10.2).
After the CC task, we also asked the participants to freely comment about their experience with the task.

We prepared a $threshold$ set to control the number of RCCs to provide.
This was determined by referring to the distribution of $\Delta^{\mathrm{w/o}}_i - \Delta^{\mathrm{w/}}_i$ calculated with the reliance dataset.

We compared the F-score for the humans' decisions between the Pred-RC condition and random condition, which was provided from the reliance dataset.
Here, the F-score was calculated with the number of times human decisions matched the AI's success.

\subsection{Hypothesis}
We hypothesized that by properly selecting the timing for providing RCCs, Pred-RC can soften the decrease in the F-score in spite of the RCCs omitted,
whereas that of the random condition was larger.

\subsection{Results}
\begin{figure}[t]
  \begin{center}
  \includegraphics[width=\linewidth]{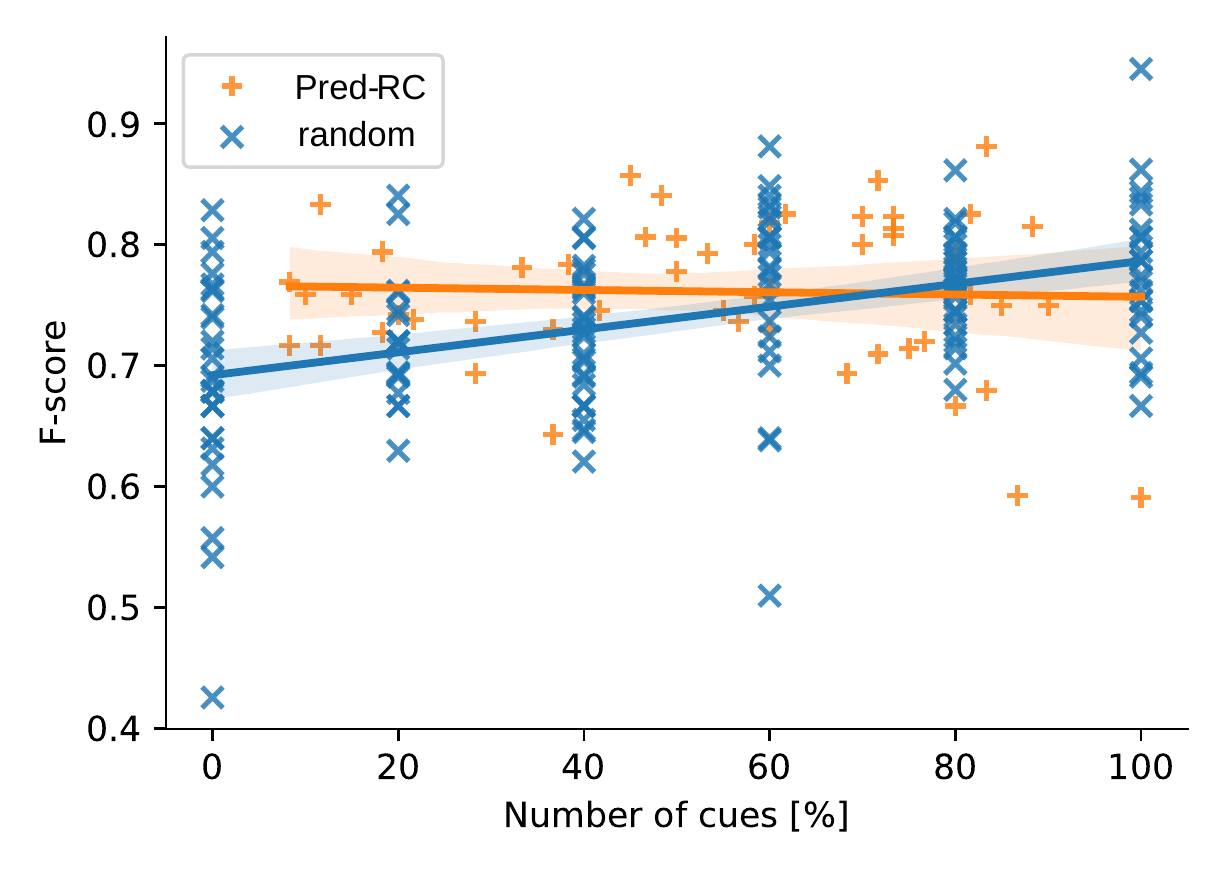}
    \caption{F-score of humans' decisions. Error bands shows 95\% CI for linear regressions.}
  \label{fig:f-score}
  \end{center}
\end{figure}

Figure~\ref{fig:f-score} illustrates the F-score for the humans' decisions.
We conducted an ANCOVA\footnote{Analysis of covariance} to statistically analyze the results.
There were significant effects for the number of cues ($F(1, 199) = 30.1; p < .0001; \eta_p^2 = .132$),
the RCC selection method ($F(1, 199) = 4.54; p = .034; \eta_p^2 = .022$),
and the interaction effect ($F(1, 199) = 7.31; p = .007; \eta_p^2 = .035$).

In the random condition, the F1-score decreased as the number of RCCs decreased.
On the other hand, as far as the range of the data, the number of RCCs had little effect on the F-score in the Pred-RC condition,
so the difference in F1-score between the Pred-RC and random conditions broadened as RCCs were reduced.
This suggests that Pred-RC softened the decrease in the F-score in spite of the omitted RCCs,
which supports our hypothesis.
Therefore, we conclude that Pred-RC can reduce the number of times RCCs are provided
while avoiding over/under-reliance by evaluating the effect of each RCC on the basis of human reliance prediction.

\section{Discussion}\label{section:discussion}
\subsection{Examples of Pred-RC's behavior}
\begin{figure}[t]
  \begin{center}
  \includegraphics[width=\linewidth]{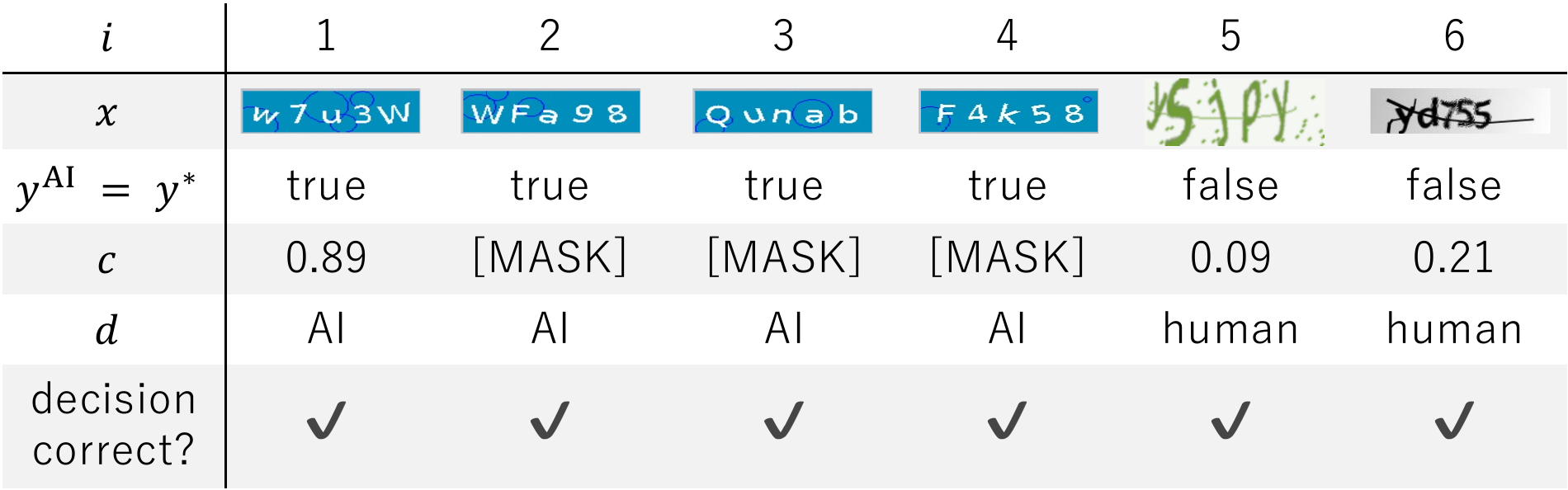}
    \caption{Successful example of Pred-RC}
  \label{fig:success}
  \end{center}
\end{figure}
Figure~\ref{fig:success} shows a successful example of Pred-RC.
We need to mention that it is difficult to follow the actual dynamics of the interaction among Pred-RC, participants, and tasks,
so our explanations here are post-hoc.
First, four images from the dataset on which the task AI achieved the best accuracy were provided, 
and actually, all the answers for them by the AI were correct.
Pred-RC decided to show a high confidence rate at the first image, and the participant could correctly assign it to the AI.
Pred-RC did not provide an RCC for the next three images, which we can consider a valid decision because they were similar to the first one,
so the participant was likely to expect the AI to be continuously successful.
As a result, the participant could properly assign them to the AI.
The fifth and sixth images were not in the datasets with which the task AI was trained, so the participant needed to do it by him/herself. 
Here, Pred-RC provided RCCs to avoid over-reliance on the AI, and the participant could avoid assigning them to the AI.

\begin{figure}[t]
  \begin{center}
  \includegraphics[width=\linewidth]{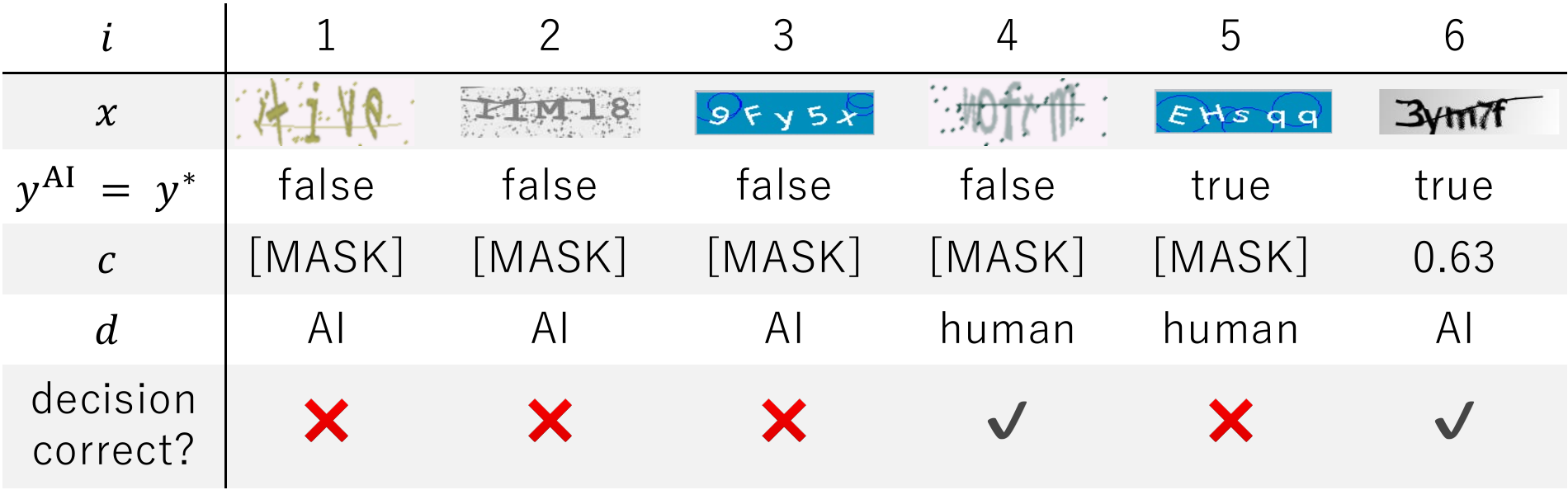}
    \caption{Unsuccessful example of Pred-RC}
  \label{fig:unsuccess}
  \end{center}
\end{figure}
Figure~\ref{fig:unsuccess} illustrates an unsuccessful example.
The participant chose the AI for the first three tasks, observed the AI's failures in a row, and eventually changed his/her decision for the fourth task.
Here, Pred-RC could not provide RCCs until the sixth task because the target percentage of RCCs was 20\%, and $threshold$ was high.
At the fifth task, we found that the reliance model predicted a high reliance rate for both with and without an RCC (96.2\% and 95.6\%, respectively),
which we can consider a false prediction because the participant had observed the failures and was likely to engage in under-reliance.
A possible reason for this result is that the reliance model overfitted the visual features of the dataset of images with a blue background color. 
Since the task AI's accuracy for the dataset was high, and the participants tended to notice this, they also tended to assign images from this dataset to the AI.
The model may have overestimated human reliance and been less sensitive to history data when images from this dataset were given.
In contrast, the model and Pred-RC made a reasonable prediction and decision at the sixth task.
The model predicted a high reliance rate with the RCC (65.7\%) and low without it (34.9\%).
Because the success probability was 85.4\%, Pred-RC decided to provide the RCC, and the participant could successfully choose the AI.

\subsection{Participants' comments}
We asked the participants to freely comment about their experiences during the CC task
and acquired comments from 45 of them.

Twenty-one participants mentioned that they had focused on specific visual features such as ``blue background color'' or ``dots'' to decide to whom to assign tasks, suggesting that they were aware of the bias of the task AI's success probability,
and most of them successfully captured the characteristics of the CAPTCHA datasets.

Six participants mentioned the provided RCCs.
While one negatively assessed them (``I felt the confidence rate was not reliable''),
the others provided positive comments (``I found that the AI was likely to succeed when the confidence rate was more than 50\%, so I chose the AI then.'').
This may indicate that expectations toward the reliability of RCCs are differ by the individual, which may arouse distrust in RCCs.
Using meta-RCCs, which calibrate trust not in an AI but RCCs, or multiple RCCs to make up for distrusted ones is promising as well.
Pred-RC can theoretically afford multiple RCCs for its input by changing $c_i$, which is a future direction for extending Pred-RC.
Depending on the situation, the problem of distrusted RCCs should be handled in another way such as apologies, excuses, or explanations and dialogues, which are found to be effective for trust repair~\cite{10.1145/3171221.3171258,10.1145/3319502.3374839,10.1007/978-3-319-25554-5_57,10.5555/3378680.3378691,10.1145/3171221.3171275}.

\subsection{Limitation}
An important limitation of Pred-RC is that it does not consider human capability for a task.
Two participants commented that they used the task AI when they were not confident in their answers.
In the CC task, humans were not perfect as well ($84.4\%$ accuracy when $d_i = \mathrm{human}$).
While our experiments successfully demonstrated that Pred-RC can effectively calibrate human reliance
with a measure of how many times humans assign a task to the AI if and only if the AI can succeed,
to improve the total collaboration performance,
we still need to take into account the capability of a human and compare it with that of an AI.

We attempted to control the number of RCCs by changing $threshold$.
However, in actual use, we need to consider the trade-off between collaboration performance and the communication cost of RCCs
rather than rigidly target the number of RCCs.
A future direction for this work is to integrate machine-learning methods to adjust the threshold.
A possible approach is using reinforcement learning (RL), in which another reliance model learns not $r_i$ but $threshold$
with a reward function that balances the performance and cost.

%% file: 6_conclusion.tex
\section{Conclusion}\label{section:conclusion}
This paper proposed Pred-RC, a method for selectively providing RCCs for human-AI collaboration.
It decides whether to provide an RCC to avoid a discrepancy between the task success probability of an AI and the human reliance rate.
In Pred-RC, a cognitive reliance model is used to predict reliance on an AI given a specific task on the basis of the collaboration history with a human.
It can predict whether a human will assign a task to an AI for cases in which an RCC is provided or not.
We implemented and tested Pred-RC for a human-AI collaboration task called the CC task.
The results demonstrated that Pred-RC can perform reliance calibration with a reduced number of RCCs,
indicating that we can selectively provide RCCs by evaluating their influence on human reliance with reliance prediction.